\documentclass{article}
\usepackage{spconf,amsmath,amsfonts,graphicx,hyperref}
\usepackage{booktabs}
\usepackage{multirow}
\usepackage{tikz}

\renewcommand{\paragraph}[1]{%
  \par\addvspace{1.2ex plus 0.3ex minus 0.2ex}%
  \noindent\textbf{#1}\hspace{0.5em}}

\title{Corrective Forcing: Unified Post-Training for Diffusions and Flows\\ in Generative Speech Enhancement}

\name{Qing Yao, Lijian Gao, and Qirong Mao$^{\dagger}$%
\thanks{$^{\dagger}$ represents the corresponding author. This work was supported in part by the National Natural Science Foundation of China (Grant Nos. 62576155, 62506144, and 62176106), the Natural Science Foundation of Jiangsu Province (Grant No. SH2025108), and the Postgraduate Research \& Practice Innovation Program of Jiangsu Province (Grant No. KYCX25\_4234). Code and demo are available at \url{https://yorch233.github.io/CoF}.}}

\address{\parbox{\textwidth}{\centering\ninept
School of Computer Science and Communication Engineering, Jiangsu University\\
Jiangsu Engineering Research Center of Big Data Ubiquitous Perception and Intelligent Agriculture Applications\\
Provincial Key Laboratory of Computational Intelligence and New Technologies in Low-Altitude Digital Agriculture\\
Zhenjiang, China}}

\begin{document}
\ninept
\maketitle
\begin{abstract}
    Diffusion and flow models, as promising generative paradigms for speech enhancement, face a training--inference mismatch: training uses analytical path states, whereas inference recursively evaluates models on self-generated rollout states along discretized sampling trajectories. This mismatch causes prediction and discretization errors to accumulate. To address it, we introduce Corrective Forcing (CoF), a post-training paradigm that forces diffusion and flow models to learn from self-generated rollouts and correct their predictions. CoF corrects clean-speech predictions on rollout states toward the ground truth under dynamic sampling schedules, exposing the model to varying inference conditions. It further regularizes local evolution using locally corrected counterfactual transitions as references for factual transitions. By expressing model outputs through a shared clean-speech prediction parameterization, CoF applies the same post-training objective across diffusion and flow formulations. Experiments with SB-VE and OT-CFM demonstrate improvements in perceptual quality and reconstruction fidelity, together with robust performance across different numbers of sampling steps.
\end{abstract}
\section{Introduction} \label{sec:intro}
Speech enhancement (SE) aims to recover clean speech from degraded recordings, and generative SE approaches model the underlying conditional distribution of clean speech \cite{richter2023speech}, achieving high perceptual naturalness in reconstructed speech. Recent studies have adapted diffusion and flow models to learn conditional stochastic or deterministic dynamics that transport between degraded and clean speech distributions. Specifically, diffusion-based SE methods \cite{richter2023speech, lemercier2023storm} define probability paths through stochastic differential equations (SDEs) and train a generative model to learn score fields for constructing reverse-time stochastic dynamics. Schr{\"o}dinger bridge (SB) \cite{jukic2024schrodinger}, formulated as entropy-regularized optimal transport, further provides a principled stochastic formulation bridging degraded and clean speech distributions, achieving high-quality enhancement with efficient sampling. In parallel, flow matching (FM) methods optimize a generative model to learn time-dependent vector fields, enabling deterministic dynamics via ordinary differential equations (ODEs) and facilitating efficient few-step enhancement \cite{lipman2023flow, le2025flowse}. 

Despite these different formulations, both families face a fundamental training--inference mismatch: training optimizes models on the analytical states sampled from prescribed probability paths, whereas inference recursively evaluates them on self-generated rollout states produced from previous predictions along a discretized sampling trajectory \cite{wang2026rethinking}. This mismatch, referred to as \textit{exposure bias} \cite{bengio2015scheduled, ning2024elucidating, ning2023input}, arises as prediction errors move rollout states away from analytical training states, inducing a progressive distribution shift and error accumulation \cite{ross2011dagger, lu2022dpmsolver}. Thus, while standard training can learn powerful generative dynamics, the model is never exposed to its own inference behavior, limiting generalization across rollout distributions induced by different sampling schedules.

To address this, Regularized Schr{\"o}dinger Bridge (RSB) perturbs training states and targets to simulate inference deviations \cite{yao2026rsb}, but does not expose the model to its actual rollout distributions. In contrast, Correcting the Reverse Process (CRP) uses empirical rollouts \cite{lay2024single}, but optimizes only the final model call under a fixed number of function evaluations (NFE), thereby restricting both the optimization scope and the rollout conditions encountered during training. In effect, these SE attempts mainly improve predictions over simulated or partial rollout states, while errors can still propagate through local transitions, i.e., numerical solver updates. Robust inference therefore requires not only accurate predictions but also transition behavior that limits error accumulation across sampling steps \cite{kim2024consistency}. Accordingly, these limitations motivate us to (1) force the models to learn from their own rollout distributions under dynamic sampling schedules \cite{lu2022dpmsolver, huang2025selfforcing}, thereby generalizing across different inference conditions; and (2) regularize their local evolution under transition-induced deviations before errors accumulate along the trajectory.

In this paper, we introduce \emph{Corrective Forcing (CoF)}, a post-training paradigm for generative SE that forces pretrained models to learn from their own rollout distributions across sampling schedules and correct predictions on them, with corrective supervision at both rollout-state and local-transition levels. First, dynamic rollout correction (DRC) corrects clean-speech predictions on rollout states under dynamic sampling schedules, thereby exposing the model to varying inference conditions. Furthermore, counterfactual transition consistency (CTC) regularizes local evolution under state deviations by aligning predictions at states reached through factual and locally corrected counterfactual transitions, with an exponential moving average (EMA) teacher providing the corrective reference. We validate CoF on Schr{\"o}dinger bridge with variance-exploding diffusion (SB-VE) and optimal transport conditional flow matching (OT-CFM), where a shared clean-speech prediction parameterization enables a unified post-training objective across formulations. Experiments demonstrate improvements in perceptual quality and reconstruction fidelity, together with generalization across different NFEs.

\section{Background} \label{sec:related}

\subsection{Schr{\"o}dinger Bridge for Speech Enhancement} 
Among diffusion-based approaches to SE, SB \cite{jukic2024schrodinger} models probability transport between paired degraded speech $y$ and clean speech $x$ as an entropy-regularized optimal transport problem \cite{liu2023i2sb}, equivalently governed by a pair of forward--backward SDEs
\begin{equation}
  \label{eq:sb_sde}
  \begin{split}
    dX_t &= [\mathbf{f}_t + g_t^2 \nabla \log \Psi_t(X_t)]\,dt + g_t\,d\mathbf{W}_t,\\
    dX_t &= [\mathbf{f}_t - g_t^2 \nabla \log \hat{\Psi}_t(X_t)]\,dt + g_t\,d\bar{\mathbf{W}}_t,
  \end{split}
\end{equation}
where $X_t \in \mathbb{C}^{F \times L}$ represents a complex short-time Fourier transform (STFT) spectrogram at time $t$ with $F$ frequency bins and $L$ time frames, where $X_0=x$ and $X_1=y$. $\mathbf{f}_t$ and $g_t$ are the drift and diffusion coefficients. $\Psi_t$ and $\hat{\Psi}_t$ denote potential functions. $\mathbf{W}_t$ and $\bar{\mathbf{W}}_t$ are Brownian motions. For SB-VE considered in this work, we use the Gaussian marginal $p_t(X_t\mid X_0,X_1)$ defined in \cite{jukic2024schrodinger}.

\paragraph{Training.} We sample $t\sim\mathcal{U}(t_{\min},1)$ and $X_t$ from the corresponding Gaussian marginal, where $t_{\min}$ denotes the minimum training time. A generative model $X_\theta$ is trained to predict $X_0$ by minimizing
\begin{equation}
  \label{eq:sb_objective}
  \mathcal{L}_{\mathrm{SB}}(\theta) = \mathbb{E}_{t,\,X_t} \left[ \left\|X_\theta(X_t,t,y)-X_0\right\|_2^2 \right].
\end{equation}

\subsection{Flow Matching for Speech Enhancement}
FM learns a time-dependent vector field that parameterizes deterministic dynamics between two endpoints through an ODE \cite{lipman2023flow, le2025flowse}:
\begin{equation}
  \label{eq:fm_ode}
  dX_t = u_t(X_t\mid x,y)\,dt,
\end{equation}
where $u_t$ denotes the vector field. In this work, we adopt the
OT-CFM formulation used in FlowSE \cite{le2025flowse} and construct
the Gaussian path $X_t=(1-t)X_0+tX_1$, where $X_0=x$,
$X_1=y+\sigma_{\max}\mathbf{z}$, and
$\mathbf{z}\sim\mathcal{N}_{\mathbb{C}}(\mathbf{0},\mathbf{I})$. $\sigma_{\max}$ specifies the Gaussian perturbation scale at the degraded endpoint. The corresponding vector field is
\begin{equation}
u_t(X_t \mid x, y)=X_1 - X_0.
\end{equation}

\paragraph{Training.} We sample $t\sim\mathcal{U}(t_{\min},1)$ and $X_t$ from the Gaussian path. A generative model $v_\theta$ is trained to predict $u_t$ by minimizing
\begin{equation}
  \label{eq:fm_objective}
  \mathcal{L}_{\mathrm{FM}}(\theta) = \mathbb{E}_{t,\,X_t} \left[ \left\|v_\theta(X_t,t,y) -u_t\right\|_2^2 \right].
\end{equation}

\subsection{A Unified View of Diffusions and Flows} Despite their distinct formulations, diffusion and flow models can be described under a common framework of Gaussian probability paths \cite{wang2026rethinking}. Specifically, for a formulation $\mathcal{G}$ and paired data $(x, y)$, the intermediate state $X_t$ can be sampled from the Gaussian marginal $p_t^{\mathcal{G}}(X_t\mid x, y)$ through the reparameterization
\begin{equation}
  \label{eq:interpolant}
  X_t=\alpha^\mathcal{G}_t x+\beta^\mathcal{G}_t y+\gamma^\mathcal{G}_t\mathbf{z},\qquad \mathbf{z}\sim\mathcal{N}_{\mathbb{C}}(\mathbf{0},\mathbf{I}),
\end{equation}
where $(\alpha_t^\mathcal{G},\beta_t^\mathcal{G},\gamma_t^\mathcal{G})$ are the formulation-specific coefficients.

Under this framework, standard training optimizes a generative model $G_\theta$ on analytical states sampled from $p_t^{\mathcal{G}}$ to predict clean speech, scores, or vector fields. During inference, a formulation-specific solver evaluates the model recursively along a discretized schedule $\mathcal T_N^{\mathcal G}=\{t_i\}_{i=0}^{N}$, where
\begin{equation}
  1=t_0>t_1>\cdots>t_{N-1}>t_{N}=0.
\end{equation}
Here, $N$ denotes the number of sampling steps, which equals the NFE during inference for both SB-VE and OT-CFM. Both formulations start from the endpoint state at $t_0=1$ and recursively move the state from the degraded endpoint toward the clean endpoint, although SB-VE  usually uses stochastic SDE updates defined in \cite{jukic2024schrodinger} whereas OT-CFM uses deterministic ODE updates defined in \cite{le2025flowse}. This recursive evaluation induces an empirical rollout distribution $q_t^{\mathcal{G},N}$ that can deviate from the analytical training distribution $p_t^{\mathcal{G}}$. Consequently, this mismatch exposes a limitation of conventional training: the model is optimized on analytical states but deployed on states generated by its own numerical solver.

\section{Corrective Forcing} \label{sec:method} 
To address the training--inference mismatch, we formalize generative SE as a two-stage paradigm: pre-training establishes foundational generative dynamics on analytical distributions, while \emph{CoF} adapts the pretrained models for inference on empirical rollout states.

\subsection{From Pre-Training to Post-Training} \label{ssec:two_stage}

\paragraph{Pre-Training on Analytical Distributions.} During pre-training, $G_\theta$ is optimized on analytical states sampled from $p_t^{\mathcal G}$ to learn the formulation-specific generative dynamics. Specifically, pre-training of SB-VE \cite{jukic2024schrodinger} utilizes the data-prediction loss in Eq.~\eqref{eq:sb_objective}, and that of OT-CFM \cite{le2025flowse} utilizes the vector-field prediction loss in Eq.~\eqref{eq:fm_objective}.

\paragraph{Post-Training on Rollout Distributions.} During post-training, $G_\theta$ is adapted directly to its empirical rollout distributions. We first define the \emph{one-step transition} from time $t_i$ to $t_{i+1}$ as
\begin{equation}
  \label{eq:transition}
  \mathcal{S}_{\theta,t_i\rightarrow t_{i+1}}^{\mathcal{G}}(X_{t_i})=\Phi^{\mathcal{G}}\!\left(X_{t_i},G_\theta(X_{t_i},t_i,y),t_i,t_{i+1};y,\xi\right),
\end{equation}
where $\Phi^{\mathcal{G}}$ is the solver update rule of $\mathcal{G}$. Here, $\xi\sim\mathcal{N}_{\mathbb{C}}(\mathbf{0},\mathbf{I})$ denotes the Gaussian noise used by an SDE solver and is omitted for an ODE solver. We retain it as an implicit argument of $\mathcal S$. During inference, given an inference schedule $\mathcal{T}_N^{\mathcal{G}}=\{t_i\}_{i=0}^N$, its corresponding rollout states form a sampling trajectory, i.e., $\{\tilde{X}_{t_k}\}_{k=0}^{N}$. This trajectory is recursively generated by the one-step transitions as
\begin{equation}
  \label{eq:rollout}
  \tilde{X}_{t_{k+1}}=\mathcal{S}_{\theta,t_k\rightarrow t_{k+1}}^{\mathcal{G}}\!\left(\tilde{X}_{t_k}\right),\qquad k=0,\ldots,N-1,
\end{equation}
where $\tilde{X}_{t_0}=X_1$ and $\tilde{X}_{t_N}$ is the final estimate of $X_0$. The rollout procedure induces an empirical state distribution $q_t^{\mathcal G,N}$ at each time $t$, from which states are drawn during post-training.

\subsection{Dynamic Rollout Correction (DRC)} 
Given the rollout distribution defined above, DRC directly corrects the model prediction at each self-generated rollout state toward the ground-truth clean-speech endpoint. Rather than forcing rollout states back onto the prescribed probability path, DRC encourages predictions along the rollout to remain consistent with the clean-speech target, facilitating reconstruction faithfulness.

\paragraph{Rollout Generation on Dynamic Schedules.} Different numbers of sampling steps produce different transition intervals and consequently distinct rollout distributions. Therefore, DRC exposes the model to rollout states under varying sampling schedules, enabling it to generalize across different inference conditions.

Specifically, at each post-training iteration, we uniformly sample $N_d \sim \mathcal{U}\{1,\dots,N_{\max}\}$ and construct the corresponding sampling schedule $\mathcal{T}_{N_d}^{\mathcal{G}}$ following the standard discretization adopted by formulation $\mathcal{G}$ \cite{jukic2024schrodinger, le2025flowse}. Here, varying $N_d$ across iterations induces a dynamic family of rollout distributions. We then uniformly sample an index $i\in\{0,\dots,N_d-1\}$, corresponding to time $t_i$ in $\mathcal{T}_{N_d}^{\mathcal{G}}$. Starting from $\tilde X_{t_0}=X_1$, we apply the one-step transitions in Eq.~\eqref{eq:rollout} for $i$ steps to obtain the rollout state $\tilde X_{t_i}\sim q_{t_i}^{\mathcal{G},N_d}$.

\paragraph{Unified Clean-Speech Prediction for Correction.} To apply DRC across formulations, we express their original model outputs through a common clean-speech prediction parameterization. Specifically, the output of $G_\theta$ is mapped to a clean-speech estimate $\hat X_\theta(\tilde X_t,t,y)$:
\begin{itemize}
  \item \emph{Data prediction (SB-VE):} $G_\theta=X_\theta$ directly predicts $X_0$, so the clean-speech estimate is
  $\hat X_\theta(\tilde X_t,t,y)=G_\theta(\tilde X_t,t,y)$.
  
    \item \emph{Vector-field prediction (OT-CFM):}
    $G_\theta=v_\theta$ predicts the vector field. Since $X_t=X_0+t\,u_t$, the estimate is given by
    \begin{equation}
      \label{eq:reparam}
      \hat X_\theta(\tilde X_t,t,y)
      =\tilde X_t-t\,G_\theta(\tilde X_t,t,y).
    \end{equation}
\end{itemize}
Beyond these, the clean-speech prediction parameterization can also be extended to score or noise predictions \cite{lu2022dpmsolver}.

With this shared parameterization, we introduce a reconstruction loss $\ell_{\mathrm{rec}}$ to supervise clean-speech predictions on rollout states, aiming to preserve both spectral accuracy and waveform fidelity. Accordingly, DRC is formulated as
\begin{equation}
\label{eq:drc_loss}
\mathcal{L}_{\mathrm{DRC}}=\mathbb{E}_{N_d,t_i,\tilde{X}_{t_i}}\left[\ell_{\mathrm{rec}}\left(\hat X_\theta(\operatorname{sg}[\tilde{X}_{t_i}],t_i,y),X_0\right)\right],
\end{equation}
where $\operatorname{sg}[\cdot]$ denotes the stop-gradient operator. $\ell_{\mathrm{rec}}$ is a composite reconstruction criterion following \cite{wang2026rethinking}, consisting of magnitude-spectrum mean squared error (MSE), complex-spectrum MSE, and negative scale-invariant signal-to-noise ratio computed on reconstructed waveforms, weighted by 0.7, 0.3, and 0.01, respectively. 

\subsection{Counterfactual Transition Consistency (CTC)}
Although DRC improves predictions at rollout states, residual errors may still affect subsequent model inputs through numerical transitions. To provide such transition-level supervision, CTC constructs factual and counterfactual transitions from the same rollout state, where the counterfactual branch represents a locally corrected evolution and provides a corrective reference for the factual branch.  

\paragraph{Factual vs. Counterfactual Transitions.} To stabilize the factual transition and provide a reliable counterfactual reference, we maintain an EMA teacher $G_{\bar\theta}$~\cite{tarvainen2017mean}, initialized from the pretrained model as $\bar\theta\leftarrow\theta$ and updated after each post-training step via $\bar\theta\leftarrow\mu\bar\theta+(1-\mu)\theta$, where $\mu$ denotes the EMA decay rate.

Starting from $\tilde X_{t_i}$, we sample a continuous time $s\sim\mathcal{U}(t_{\min},t_i)$ and construct the \emph{factual} transition to capture the local evolution during inference. In parallel, the \emph{counterfactual} transition starts from the same state over the same time interval, but replaces the factual model prediction with the ground-truth counterpart $G_{\mathrm{GT}}^{\mathcal G}$. This yields a locally corrected evolution toward the clean-speech endpoint. The resulting factual and counterfactual states are given by
\begin{equation}
  \label{eq:ctc_branches}
  \begin{split}
    \tilde X_s^{\mathrm{F}}
    &=\Phi^{\mathcal G}\!\left(
    \tilde X_{t_i},
    G_{\bar\theta}(\tilde X_{t_i},t_i,y),
    t_i,s;y,\xi
    \right),\\
    \tilde X_s^{\mathrm{CF}}
    &=\Phi^{\mathcal G}\!\left(
    \tilde X_{t_i},
    G_{\mathrm{GT}}^{\mathcal G},
    t_i,s;y,\xi
    \right),
  \end{split}
\end{equation}
with shared $\xi$. Here, $G_{\mathrm{GT}}^{\mathcal G}$ specifies the corrective intervention on the factual prediction, given by $X_0$ for SB-VE and $(\tilde X_{t_i}-X_0)/t_i$ for OT-CFM, which is obtained by rearranging Eq.~\eqref{eq:reparam}.

\paragraph{Transition Consistency as Regularization.} Given the paired transition states, CTC uses the teacher prediction at the counterfactual state as a corrective reference for the online prediction at the factual state. Specifically, the online prediction is $\hat X_{0,\mathrm{F}}^s=\hat X_\theta(\operatorname{sg}[\tilde X_s^{\mathrm{F}}],s,y)$, while the counterfactual target is $\hat X_{0,\mathrm{CF}}^s=\operatorname{sg}[\hat X_{\bar\theta}(\tilde X_s^{\mathrm{CF}},s,y)]$. The consistency objective is then:
\begin{equation}
  \label{eq:ctc_loss}
  \mathcal{L}_{\mathrm{CTC}}=\mathbb{E}_{N_d,t_i,\tilde X_{t_i},s,\xi}\!\left[\ell_{\mathrm{rec}}\!\left(\hat X_{0,\mathrm{F}}^s,\hat X_{0,\mathrm{CF}}^s\right)\right].
\end{equation}
Here, CTC acts as a regularizer that encourages consistent predictions under factual and locally corrected counterfactual transitions, aiming to reduce sensitivity to transition-induced state deviations.

\subsection{Overall Objectives}

Overall, the post-training objective is defined as
\begin{equation}
  \label{eq:cof_loss}
  \mathcal{L}_{\mathrm{CoF}}=\mathcal{L}_{\mathrm{DRC}}+\lambda_{\mathrm{CTC}}\mathcal{L}_{\mathrm{CTC}}.
\end{equation}

\begin{figure*}[t!]
  \centering
  \includegraphics[width=1 \textwidth]{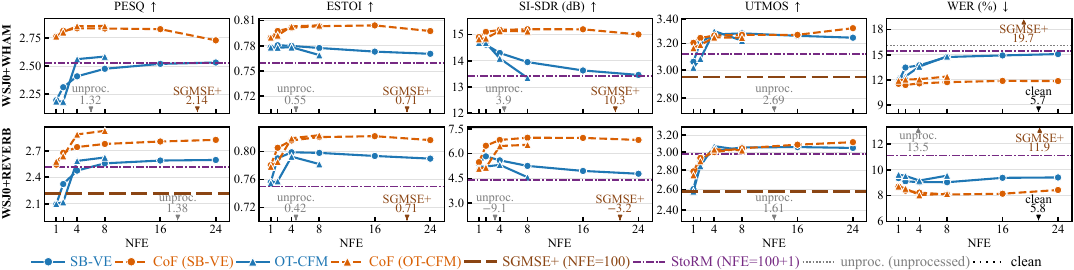}
  \vspace{-8mm}
  \caption{Performance comparison with and without CoF at different NFEs on WSJ0+WHAM (top) and WSJ0+REVERB (bottom). }
  \vspace{-5mm}
  \label{fig:nfe_curves}
\end{figure*}

\begin{table}[t]
  \centering
  \fontsize{9}{10}\selectfont
  \renewcommand{\arraystretch}{0.92}
  \setlength{\tabcolsep}{1pt}
  \setlength{\aboverulesep}{1pt}
  \setlength{\belowrulesep}{1.5pt}
  \begin{tabular*}{\linewidth}{@{\extracolsep{\fill}}lc|cccc@{}}
    \toprule
    Method                                           & NFE       & PESQ$\uparrow$ & ESTOI$\uparrow$ & SI-SDR$\uparrow$ & UTMOS$\uparrow$\\
    \midrule
    Unprocessed                                      & --        & 1.97           & 0.79            & 8.5              & 2.85 \\
    SEMamba \cite{chao2024semamba}                   & 1         & 3.54           & \textbf{0.89}   & 19.7             & 3.52 \\
    MP-SENet \cite{lu2025mpsenet}                    & 1         & \textbf{3.61}  & \textbf{0.89}   & 19.4             & 3.52 \\
    \midrule
    SGMSE+ \cite{richter2023speech}                  & 100       & 2.92           & 0.86            & 17.5             & \textbf{3.68} \\
    StoRM \cite{lemercier2023storm}                  & $100{+}1$ & 2.93           & 0.87            & 18.6             & 3.54 \\
    RSB \cite{yao2026rsb}                            & 50        & 3.04           & 0.87            & 18.8             & 3.64 \\
    CRP \cite{lay2024single}                         & 5         & 3.08           & 0.88            & 19.3             & -- \\
    \midrule
    \multirow{3}{*}{SB-VE \cite{jukic2024schrodinger}} & 1         & 2.96           & 0.88            & 19.7             & 3.43 \\
    & 4         & 3.07           & 0.88            & 18.9             & 3.60 \\
    & 16        & 3.08           & 0.87            & 18.1             & 3.63 \\
    \midrule
    \multirow{3}{*}{CoF (SB-VE)}                       & 1         & 3.17           & 0.88            & 19.4             & 3.58 \\
    & 4         & 3.21           & 0.88            & 19.4             & 3.63 \\
    & 16        & 3.21           & 0.88            & 19.4             & 3.63 \\
    \midrule
    \multirow{2}{*}{OT-CFM \cite{le2025flowse}}     & 1         & 2.89           & 0.87            & 19.7             & 3.46 \\
    & 4         & 3.08           & 0.87            & 19.1             & 3.61 \\
    \midrule
    \multirow{2}{*}{CoF (OT-CFM)}                   & 1         & 3.07           & 0.88            & 19.7             & 3.61 \\
    & 4         & 3.18           & 0.88            & \textbf{19.8}    & 3.63 \\
    \bottomrule
  \end{tabular*}
  \caption{Performance comparison on VoiceBank+DEMAND.}
  \label{tab:vb_demand}
\end{table}

\section{Experiments} \label{sec:experiments}
\subsection{Experimental Setup} \label{ssec:setup}

\paragraph{Datasets.} We evaluate denoising performance on the VoiceBank+DEMAND \cite{valentini2016investigating} and WSJ0+WHAM datasets, and dereverberation performance on the WSJ0+REVERB dataset. Specifically, WSJ0+WHAM is constructed by mixing clean WSJ0 speech \cite{garofolo1993wsj0} with WHAM noise \cite{wichern2019wham}, while WSJ0+REVERB is generated by convolving WSJ0 utterances with simulated room impulse responses. The data splits and construction procedures for these two datasets follow those used in RSB \cite{yao2026rsb}. Additionally, VoiceBank+DEMAND is a publicly available benchmark widely used for speech denoising. For all datasets, audio is sampled at 16~kHz and converted into complex spectrograms using an STFT with a square-root Hann window of length 510 and a hop size of 128. We then apply square-root magnitude warping to the complex spectrograms.

\paragraph{Evaluation Metrics.} We assess performance with reference-based and reference-free metrics. Specifically, the reference-based metrics include PESQ \cite{rix2001perceptual} for perceptual quality, ESTOI \cite{jensen2016algorithm} for speech intelligibility, SI-SDR \cite{leroux2019sdr} for signal fidelity, and the corpus-level word error rate (WER) of transcriptions generated by Whisper base.en \cite{radford2023whisper}. We further employ the reference-free UTMOSv2 (denoted as UTMOS) from VoiceMOS Challenge 2024  \cite{baba2024utmosv2} to assess overall speech perceptual quality. Best results are shown in bold.

\paragraph{Compared Baselines.} We consider SB-VE \cite{jukic2024schrodinger} and OT-CFM \cite{le2025flowse} as the baseline generative models and apply CoF to both through post-training. For broader comparison, we include representative predictive methods, namely, SEMamba \cite{chao2024semamba} and MP-SENet \cite{lu2025mpsenet}, together with generative SE methods, including SGMSE+ \cite{richter2023speech}, StoRM \cite{lemercier2023storm}, RSB \cite{yao2026rsb}, and CRP \cite{lay2024single}. SGMSE+, StoRM, and RSB are evaluated using their recommended samplers with 50 sampling steps, corresponding to 100, 101, and 50 NFEs, respectively. 

\paragraph{Implementation Details.} On VoiceBank+DEMAND, results for the baselines are obtained by inference with their released checkpoints, except for CRP, whose results are taken from \cite{le2025flowse}. On WSJ0+WHAM and WSJ0+REVERB, we reproduce SGMSE+ and StoRM using their official code for reference. We reimplement SB-VE and OT-CFM using the same NCSN++M architecture \cite{lemercier2023storm}, following the default implementations and parameters of SB \cite{jukic2024schrodinger} and FlowSE \cite{le2025flowse}, respectively. The best pretrained checkpoint is used for CoF post-training. We set $\lambda_{\mathrm{CTC}}=0.1$ in Eq.~\eqref{eq:cof_loss} and use an EMA decay rate of $\mu=0.999$. We optimize $\mathcal{L}_{\mathrm{CoF}}$ using Adam with a learning rate of $10^{-4}$ and a total batch size of 8 on two NVIDIA RTX 5090 GPUs. We set $(N_{\max},t_{\min})$ to $(16,10^{-4})$ for SB-VE and $(8,0.03)$ for OT-CFM. For both SB-VE and OT-CFM, the sampling times in $\mathcal{T}_N^{\mathcal G}$ are uniformly spaced over the time interval. CoF uses 4,000 post-training steps and requires on average \((N_{\max}-1)/4+4\) model evaluations per step, corresponding to 7.75 for SB-VE and 5.75 for OT-CFM. We evaluate the EMA checkpoint every 200 steps on 50 validation utterances from each dataset and select the EMA checkpoint with the best validation PESQ for final inference.

\subsection{Experimental Results} \label{ssec:experiment_results}
\paragraph{Comparison with Baselines.} Table~\ref{tab:vb_demand} compares CoF with the baselines on VoiceBank+DEMAND. Overall, CoF improves PESQ for both SB-VE and OT-CFM and substantially improves multi-step SI-SDR, while generally preserving ESTOI and UTMOS. Specifically, CoF (SB-VE) reaches PESQ $3.21$ at NFE~$=4$, exceeding SB-VE at $3.07$, RSB at $3.04$, and CRP at $3.08$, while requiring fewer NFEs than the latter two. Moreover, CoF stabilizes multi-step reconstruction fidelity. In particular, CoF (SB-VE) maintains SI-SDR at $19.4$~dB across NFEs, and CoF (OT-CFM) raises SI-SDR to $19.8$~dB at NFE~$=4$. The resulting SI-SDR is also comparable to that of the predictive baselines MP-SENet and SEMamba.

Fig.~\ref{fig:nfe_curves} shows that CoF consistently improves most metrics across NFEs on the more challenging WSJ0+WHAM and WSJ0+REVERB benchmarks. At NFE~$=4$, CoF (OT-CFM) raises PESQ to $2.85$ on WSJ0+WHAM and $2.89$ on WSJ0+REVERB, with SI-SDR gains of $1.0$ and $1.1$~dB, respectively. CoF (SB-VE) also mitigates fidelity degradation under deeper rollouts. On WSJ0+WHAM, as NFE increases from $1$ to $16$, its SI-SDR improves from $14.9$ to $15.2$~dB, whereas SB-VE degrades from $14.8$ to $13.6$~dB. Overall, CoF improves PESQ and ESTOI at matched NFEs and enhances multi-step SI-SDR, while its modest and inconsistent UTMOS gains indicate that improvements in reference-free perceptual quality remain limited. CoF also maintains low WERs across NFEs, largely avoiding the degradation observed in pretrained models under deeper rollouts.

\begin{table}[t]
  \centering
  \fontsize{9}{10}\selectfont
  \renewcommand{\arraystretch}{0.92}
  \setlength{\tabcolsep}{1pt}
  \setlength{\aboverulesep}{1pt}
  \setlength{\belowrulesep}{1.5pt}
  \begin{tabular*}{\linewidth}{@{\extracolsep{\fill}}l|ccc@{}}
    \toprule
    Variants                                              & PESQ$\uparrow$ & SI-SDR$\uparrow$ & UTMOS$\uparrow$\\
    \midrule
    SB-VE pre-training                                    & 3.07           & 18.9             & 3.60 \\
    \midrule
    Fine-tuning with $\ell_{\mathrm{rec}}$                & 3.16           & 18.6             & \textbf{3.64} \\
    \midrule
    DRC                                                   & 3.19\rlap{\textsuperscript{$\ddagger$}}          & 19.2\rlap{\textsuperscript{$\ddagger$}}             & 3.61 \\
    \quad + CTC (CoF), $\lambda_{\mathrm{CTC}}=0.1$       & \textbf{3.21}\rlap{\textsuperscript{*}}   & 19.4\rlap{\textsuperscript{*}}             & 3.63\rlap{\textsuperscript{*}} \\
    \quad + CTC (CoF), $\lambda_{\mathrm{CTC}}=0.25$      & 3.20           & 19.5             & 3.63 \\
    \midrule
    CoF with consistency target $X_0$ & 3.20           & 19.3             & 3.61 \\
    CoF with fixed $N_d=8$                     & 3.16           & 19.3             & 3.62 \\
    CoF with plain MSE                                    & 3.03           & \textbf{19.6}    & 3.58 \\
    \bottomrule
  \end{tabular*}
  \parbox{\linewidth}{
    ($\ddagger$) and (*) indicate significant improvements over fine-tuning and DRC, respectively (utterance-level paired $t$-test, $p<0.05$).
    }
  \vspace{-3mm}
  \caption{Ablation study on VoiceBank+DEMAND with SB-VE.}
  \label{tab:ablation}
  \vspace{-5mm}
\end{table}

\paragraph{Ablation Study.}
Table~\ref{tab:ablation} examines the contributions of CoF with SB-VE at NFE~$=4$. Compared with fine-tuning on analytical states using $\ell_{\mathrm{rec}}$ for 4{,}000 steps, DRC significantly improves PESQ from 3.16 to 3.19 and SI-SDR from 18.6 to 19.2 dB, supporting the benefit of corrective supervision on rollout states, although UTMOS decreases from 3.64 to 3.61. Adding CTC with $\lambda_{\mathrm{CTC}}=0.1$ yields further significant improvements in PESQ, SI-SDR, and UTMOS. Increasing $\lambda_{\mathrm{CTC}}$ to 0.25 slightly favors SI-SDR but lowers PESQ. We therefore use $\lambda_{\mathrm{CTC}}=0.1$ as the default. Replacing the counterfactual teacher prediction with $X_0$ as the consistency target yields weaker overall performance, supporting the use of a transition-aware counterfactual reference rather than an additional reconstruction target. Using a fixed $N_d=8$ also degrades performance, supporting dynamic sampling schedules that expose the model to rollouts induced by diverse inference conditions. Finally, replacing the composite $\ell_{\mathrm{rec}}$ with plain MSE further raises SI-SDR to 19.6 dB but substantially reduces PESQ and UTMOS, highlighting the tradeoff introduced by the training objective. 

\section{Conclusion} \label{sec:conclusion} 
In this paper, we propose CoF to mitigate the training--inference mismatch in generative SE by adapting pretrained models to self-generated sampling trajectories. DRC corrects predictions on rollout states, while CTC regularizes local transitions through counterfactual references. Experiments show that CoF improves reference-based perceptual quality and reconstruction fidelity for both SB-VE and OT-CFM, particularly under multi-step sampling. 

\section{Compliance with ethical standards}
This study used existing speech datasets and involved no new collection of human-subject data. No ethical approval was required.

\bibliographystyle{IEEEbib}
\bibliography{refs}

\end{document}